\documentclass[11pt,a4paper]{article}
\usepackage[hyperref]{acl2017}
\usepackage{times}
\usepackage{latexsym}
\usepackage{multirow}
\usepackage{multicol}
\usepackage{lipsum}
\usepackage[T2A,T1]{fontenc}
\usepackage[utf8]{inputenc}
\usepackage[russian,english]{babel}

\usepackage{url}

\aclfinalcopy 

\title{Human Associations Help to Detect \\ Conventionalized  Multiword Expressions 
 }

\author{Natalia Loukachevitch \\
  Lomonosov Moscow State University \\
  Leniskie Gory,1\\
  Moscow, Russia \\
  {\tt louk\_nat@mail.ru} \\\And
  Anastasia Gerasimova \\
   Lomonosov Moscow State University \\
  Leniskie Gory,1 \\
   Moscow, Russia \\
  {\tt anastasiagerasimova432@gmail.com} \\}

\date{}

\begin{document}
\maketitle
\begin{abstract}
In this paper we show that if we want to obtain human evidence about conventionalization of some phrases, we should ask native speakers about associations they have to a given phrase and its component words. We have shown that if component words of a phrase have each other as frequent associations, then this phrase can be considered as conventionalized. Another type of conventionalized phrases can be revealed using two factors: low entropy of phrase associations and low intersection of component word and phrase associations. The association experiments were performed for the Russian language.
\end{abstract}

\section{Introduction}

A lot of approaches have been proposed for automatic extraction of idioms, collocations, or multiword terms from texts as potential candidates for inclusion in lexical or terminological resources \cite {Bonial,  Gelbukh, Pecina, Piasecki}.

However, developers of computational resources need clear guidelines for the introduction of phrases into their resources. Special instructions on introducing multiword terms exist for constructing information-retrieval thesauri \cite {Z39}. Developers of WordNet-like thesauri, a very popular type of resources, discuss the problem of introducing multiword expressions in their resources in several works \cite { Maziarz,Piasecki,Vincze}.
For example, it is supposed that wordnets have to include only lexicalized concepts as synsets \cite{Miller}. However, Agirre et al., \shortcite {Agirre} stress that boundaries of lexicalization are very difficult to draw. Bentivogli and Pianta \shortcite{Bentivogli} argue that there is a necessity to include non-lexicalized phrases into wordnets.

Multiword expressions comprise a broad scope of phrases including   idiomatic  expressions,  noun compounds,  technical terms, proper names, verb-particle and light verb constructions, conventionalized phrases, and others \cite {Calzolari, Sag, Baldwin}. For some of these constructions, such as idioms, it is evident that they should be included in computational lexicons. But for many of other expressions, for example,
conventionalized phrases, it is not easy to make a decision about the necessity of their inclusion. To distinguish a  multiword expression, it is important to analyze if
it has any "idiosincrasies", which can be lexical, syntactical, semantical or statistical.   

Conventionalized phrases have statistical idiosyncrasy and usually only one approach is proposed in literature to distinguish such phrases from other compositional phrases. This is so-called substitutionability test, which shows if the phrase components can be easily substituted with their synonyms \cite {Sag,  Farahmand, Farahmand1, Pearce, Senaldi}.

In this paper, we show that there are at least two more types of statistical idiosyncrasy (and related tests) to distinguish conventionalized expressions:
\begin{itemize}
\item association idiosyncrasy when components of a phrase are highly associated with each other, and 
\item relational idiosyncrasy when a phrase has lexical associations that significantly differ from the associations of its component words; usually it means that the phrase denotes a specific entity or process
with a set of its own properties and relations. 
\end{itemize}

 We provde evidence for these types of phrase idiosyncrasy in association experiments in Russian,
in which we asked  Russian native speakers what  associations they had for phrases and their component words. 
 We have found that  the human association experiment is a very efficient tool to detect conventionalized phrases with high accuracy. To the best of our knowledge, this is the first attempt to use human associations for distinguishing conventionalized phrases.

The structure of the paper is as follows. In Section 2 we consider types of phrase idiosyncrasy.  Section 3 describes the specificity of RuThes thesaurus,  from which we take phrases for the experiments. Section 4 presents the association experiment and its results. In Section 5 we test  embedding models on their capability to distinguish conventionalized phrases. Section 6 reviews related work concerning approaches of annotating 
compositionality/non-compositionality/conventionalization of noun phrases.

\section{Types of Idiosyncrasy of Multiword Expressions}
Multiword expressions are phrases that have some specificity (idiosyncrasy). Because of this, it is useful to collect them and store in lexicons and thesauri \cite {Calzolari, Sag, Baldwin}.

The idiosyncrasy can be lexical when a component of a phrase appears only within this phrase \cite {Baldwin}.
It can be syntactical when the syntactic behavior of a phrase differs from usual (for example, fixed word order).
Semantical idiosynrasy can be revealed when the meaning of a phrase cannot be inferred from
the meanings of its components. If a phrase has one of the above-mentioned types of idiosynrasy
it can be called a lexicalized expression \cite {Sag,Baldwin}. 

Statistical idiosyncrasy presupposes that the components of a phrase
co-occur more often than expected by chance. Besides, the frequency of phrases with statistical
idiosynrasy is much higher than the frequency of the phrase with one component changed to its near-synonym  (\textit {weather
forecast} vs. \textit {weather prediction}), as the result of the  substitutionability test \cite {Sag,  Farahmand1}. 
Phrases with statistical idiosyncrasy (often called   \textit {conventionalized phrases}) can be
syntactically and semantically compositional. 

In many cases conventionalized phrases are difficult to distinguish. 
For example, one  of the often mentioned conventionalized phrase \textit {traffic lights} looks fully compositional.  However, if we examine the meaning of this phrase, we can see that 
the denoted entity can be categorized as a road facility; it has signals; it is usually 
constructed on road intersections; it is needed for regulating road traffic, etc. This 
means that the phrase \textit {traffic lights} has  thesaurus relations with the correponding 
words (facilities, road, signals, regulation) that cannot be inferred from the meanings of 
its component words \textit {traffic} and  \textit {lights}.

A lot of similar examples can be found. Compositional \textit {seat belt} has  relation
to the safety concept. Food courts are usually located in shopping centers, and therefore compositional phrase \textit {food court} has  relation with the   \textit {shopping center} concept,  etc. These relations can be very useful in such NLP applications as textual entailment.

Thus, we can suppose that 
conventionalized  phrases have not only statistical idiosyncrasy, 
but also 
 \textit {relational idiosyncrasy}, which can be revealed  easier than using the substitutionability test.
The same idiosynrasy can be found in unclear cases of possible lexicalized expessions.

 In \cite {Melchuk} so-called quasi-idioms are discussed. According to Mel'{\v{c}}uk, a phrase \textit {AB} is  a quasi-idiom or weak idiom iff its meaning:  1)  includes  the  meaning  of both of its lexical components, neither as the semantic pivot, and 2) includes an additional meaning \textit {C} as its semantic pivot. Mel'{\v{c}}uk \shortcite{Melchuk} gives an example of \textit {barbed wire}, which is an obstacle, but neither \textit {barbed} nor \textit {wire} are obstacles.  Thus, it seems than \textit {semantic pivot} in this case is the  hypernym relation, that cannot be inferred from the phrase component words. It means that the quasi-idiom  is a subtype of relational idiosyncrasy.

In this paper we show that this relational idiosynrasy can be found in 
association experiments with native speakers.    Besides,  we can also reveal the association idiosyncrasy of conventionalized phrases in these experiments.

\section{RuThes Thesaurus as a Source of Conventionalized Expressions}

For the present work, we  utilized multiword expressions included in the Russian-language thesaurus  RuThes\footnote{http://www.labinform.ru/pub/ruthes/index\_eng.htm} \cite {Loukachevitch2014}. The  RuThes  thesaurus is a linguistic ontology for natural language processing, i.e. an ontology, where the majority of concepts are introduced on the basis of actual language expressions.  

RuThes has considerable similarities with WordNet: the inclusion of concepts based on senses of real text units, representation of lexical senses, detailed coverage of word senses. At the same time, the differences include attachment of different parts of speech to the same concepts, formulating  names of concepts, attention to multiword expressions,  the set of conceptual relations, etc. 

In particular, the developers of the RuThes thesaurus   have special rules for including phrases that appear compositional  into the  thesaurus. Such phrases are introduced if they have specificity in relations with other single words and/or expressions \cite {Loukachevitch2016}. The following subtypes of these expressions can be considered:
\begin{itemize}

\item A phrase is a synonym to a single word; for example, \begin{otherlanguage*}{russian} земельный участок  \end{otherlanguage*} (landing lot) is a synonym to word \begin{otherlanguage*}{russian}земля \end{otherlanguage*} (land), or a phrase has a frequent abbreviation:   \begin{otherlanguage*}{russian} заработная плата -- зарплата \end{otherlanguage*} (employee wages);

\item A phrase has a synonymous phrase and this fact cannot be simply inferred from the components of the phrase: \begin{otherlanguage*}{russian} мобильный телефон \end{otherlanguage*} (mobile phone) -- \begin{otherlanguage*}{russian} сотовый телефон \end{otherlanguage*} (cell phone);

\item A phrase generalizes several single words. Such phrases as \begin{otherlanguage*}{russian} транспортное происшествие \end{otherlanguage*} (transport accident) or \begin{otherlanguage*}{russian} учебное заведение \end{otherlanguage*} (educational institution) often look compositional but they have a very important  function of knowledge representation: they gather together similar concepts;

\item A phrase has  relations that do not follow from its component words. 
For example, the compositional phrase \begin{otherlanguage*}{russian} дорожное движение  \end{otherlanguage*} (\textit {road traffic})  has numerous relations with other phrases that cannot be inferred from its components, for example, hyponyms (\textit {left-hand traffic, one-way traffic}), related concepts (\textit {car accident, traffic jam}), etc.

\end{itemize}

Thus, phrases from RuThes without evident non-compositionality were selected for the association experiment in order to understand correlations between choice of phrases made by experts and associations of native speakers.

\section{Association Experiment}

For the experiment, we took two-word noun phrases (\textit {Adjective + Noun} and \textit {Noun + Noun-in-Genitive})  that  have high frequency in Russian newswire text collections. 

The multiword expressions  were of two main groups. The first group (Thesaurus group)  included multiword expressions from the  RuThes thesaurus. We chose phrases that either look fully compositional (\textit {increase of prices}) or that have one of components is used in a known (=described in dictionaries) metaphoric sense. This group contained 15 phrases. Another group  of phrases comprised fully compositional noun phrases not included in the thesaurus, for example, \textit {end  of January, mighty earthquake, result of work}, etc. The non-thesaurus group contained 36 phrases.

We asked respondents (mainly university students) to think of single-word associations to noun phrases. In a separate experiment, we collected associations to the component words of the same phrases. We wanted to understand if the collected associations can serve as a base for distinguishing  thesaurus phrases from non-thesaurus phrases (and as a consequence, conventionalized phrases from non-conventionalized). Twenty six native speakers gave their associations for the thesaurus phrases and twenty nine respondents participated in the experiment with non-thesaurus phrases. Forty seven people gave associations for single words.

The study was conducted via Google Forms. The respondents were asked to provide single-word associations. 
However, some participants could think only of multiword expressions. 
Such associations were also taken into account. 
Table 1 contains examples of obtained associations and their frequencies for some thesaurus phrases.

From the associations obtained, we calculated the following characteristics (Tables 2, 3):
\begin {itemize}
\item entropy of answers for single words and phrases (currently, only entropy of phrase associations was found useful and included in the tables);
\item  intersection between associations of component words and phrase associations (columns Ph1 and  Ph2 in Tables 2, 3); and
\item number of times when one component word served as an association of another component word (columns A12 and  A21 in Tables 2, 3).
\end {itemize}

Table 2 contains the results for the thesaurus phrases, and Table 3  shows partial results for the non-thesaurus phrases. 

 We can see that for thesaurus phrases, the components are associated with each other more often than for non-thesaurus phrases. The average value of such associations for thesaurus phrases is 10 times  greater than for non-thesaurus phrases. For some thesaurus phrases, both components are highly connected with another component. Withing non-thesaurus phrases, such frequent mutual associations were not found. 

\begin{table}[h]
\begin {small}
\begin{center}
\begin{tabular}{|l|l|c|}
\hline
\bf source& \bf associations&\bf freq\\
\hline
w1:\begin{otherlanguage*}{russian} земельный\end{otherlanguage*} &\begin{otherlanguage*}{russian}участок (lot)\end{otherlanguage*} &\bf 38\\
(landing)&\begin{otherlanguage*}{russian}вопрос (issue)\end{otherlanguage*} &2\\
 \hline
w2: \begin{otherlanguage*}{russian}участок\end{otherlanguage*} (lot)&\begin{otherlanguage*}{russian}земля (land)\end{otherlanguage*}&\bf 11\\
  &\begin{otherlanguage*}{russian}дача (dacha)\end{otherlanguage*}&\bf 11\\
 &\begin{otherlanguage*}{russian}полицейский\end{otherlanguage*}&4\\
&(police)&\\
&\begin{otherlanguage*}{russian}дорога (road)\end{otherlanguage*}&3\\

&\begin{otherlanguage*}{russian}дом (house)\end{otherlanguage*}&\bf 2\\
\hline
phrase:\begin{otherlanguage*}{russian}  земельный \end{otherlanguage*} &\begin{otherlanguage*}{russian}дача (dacha)\end{otherlanguage*}&\bf 12\\
\begin{otherlanguage*}{russian} участок\end{otherlanguage*}
 &\begin{otherlanguage*}{russian}дом (house)\end{otherlanguage*}&\bf 2\\
(landing lot)&\begin{otherlanguage*}{russian}надел \end{otherlanguage*}&2\\
&(allotment)&\\
 \hline
\hline
w1:\begin{otherlanguage*}{russian} повышение\end{otherlanguage*} &\begin{otherlanguage*}{russian}должность\end{otherlanguage*} & 8 \\
(increase)&(post) &\\
&\begin{otherlanguage*}{russian}зарплата\end{otherlanguage*} & 7 \\
&(wages) &\\
&\begin{otherlanguage*}{russian}работа\end{otherlanguage*} (job)& 6 \\

 \hline
w2: \begin{otherlanguage*}{russian}цена\end{otherlanguage*} (price)&\begin{otherlanguage*}{russian}ценник\end{otherlanguage*}&5 \\
  &\begin{otherlanguage*}{russian}(price-tag)\end{otherlanguage*}& \\
 &\begin{otherlanguage*}{russian}стоимость\end{otherlanguage*}&4\\
& (cost)&\\
&\begin{otherlanguage*}{russian}высокая\end{otherlanguage*}&3\\
& (high)&\\
&\begin{otherlanguage*}{russian}качество\end{otherlanguage*}&2 \\
& (quality)&\\
\hline
phrase: &\begin{otherlanguage*}{russian}инфляция\end{otherlanguage*}&\bf12 \\

\begin{otherlanguage*}{russian} \end{otherlanguage*}
\begin{otherlanguage*}{russian}повышение цен \end{otherlanguage*} &(inflation)&\\
(increase of prices)&\begin{otherlanguage*}{russian}кризис\end{otherlanguage*} (crisis)&\bf 5\\

&\begin{otherlanguage*}{russian}нефть\end{otherlanguage*} (oil)&2 \\
 \hline
\end{tabular}
\end{center}

\caption{\label{font-table} Examples of the most frequent associations for thesaurus phrases and its components}
\end {small}
\end{table}

\begin{table}[h]
\begin {scriptsize}
\begin{center}
\begin{tabular}{|l|c|c|c|c|c|}

\hline \bf Phrase &  \bf A12&\bf A21&\bf Ph1&\bf Ph2&\bf  Entr\\ \hline
\begin{otherlanguage*}{russian}транспортное \end{otherlanguage*}&  &&&&\\
\begin{otherlanguage*}{russian}происшествие\end{otherlanguage*}& 0   &7&0&6&2.16\\ 
(transport accident) &    &&&&\\\hline

\begin{otherlanguage*}{russian}учебное заведение\end{otherlanguage*}&  1  &8&\bf 1&\bf 1&\bf 2.52 \\ 
(education institute) &&&&& \\\hline
  \begin{otherlanguage*}{russian}программное \end{otherlanguage*}  &   &&&&\\ 
 \begin{otherlanguage*}{russian}обеспечение\end{otherlanguage*}  & \bf 13   &\bf 14&6&1&2.77\\
(software program)&   &&&&\\
 \hline
\begin{otherlanguage*}{russian}повышение цен\end{otherlanguage*}&0   &0&\bf 0&\bf 0&\bf 2.85  \\
(increase in prices) &   &&&&\\\hline
  \begin{otherlanguage*}{russian}земельный участок\end{otherlanguage*}  & \bf  38  & \bf  13&0&14&2.89\\ 
(landing lot) &   &&&&\\\hline
 \begin{otherlanguage*}{russian}квадратный метр\end{otherlanguage*}& \bf 10
  &\bf 20&0&1&3.22\\ 
(square meter) &   &&&&\\\hline
  \begin{otherlanguage*}{russian}электронная почта\end{otherlanguage*}& \bf 6  &\bf 12&3&4&3.27\\ 
(electronic mail) &   &&&&\\\hline

 \begin{otherlanguage*}{russian}дорожное движение\end{otherlanguage*}&0   &2&\bf 0&\bf 0&\bf 3.33\\ 
(road traffic) &   &&&&\\\hline
 \begin{otherlanguage*}{russian}заработная плата\end{otherlanguage*}  &\bf  18  &\bf 10&8&2&3.42\\ 
(employee wage) &   &&&&\\\hline
  \begin{otherlanguage*}{russian}главный герой\end{otherlanguage*}&   5 &1&\bf 0&\bf 3&\bf 3.56\\ 
(main hero) &   &&&&\\\hline
\begin{otherlanguage*}{russian}медицинская помощь\end{otherlanguage*}& 0   &5&\bf 0&\bf 4 & \bf 3.58\\ 
(medical aid) &   &&&&\\\hline
   \begin{otherlanguage*}{russian}торговый центр\end{otherlanguage*}  & \bf 26 &0&1&0&3.62 \\ 
(shopping center) &   &&&&\\\hline
 \begin{otherlanguage*}{russian}лента новостей\end{otherlanguage*}  &  \bf 18 &0&1&1&3.79 \\
(news feed) &   &&&&\\ \hline
   \begin{otherlanguage*}{russian}мобильный телефон\end{otherlanguage*}& \bf  26  &\bf 12&3&7&3.81\\ 
(mobile phone) &   &&&&\\\hline
   \begin{otherlanguage*}{russian}температура\end{otherlanguage*}&   &&&&\\ 
\begin{otherlanguage*}{russian}воздуха\end{otherlanguage*}& 4 &0&6&1&3.81\\
(air temperature) &   &&&&\\\hline

 Average&  \bf 11 &\bf 6.27&\bf 1.93&\bf 2.93&\bf 3.24  \\ \hline

\end{tabular}
\end{center}
\end {scriptsize}
\caption{\label{font-table} Results of association experiments  for the thesaurus phrases}
\end{table}

Therefore, we think that mutual associations between phrase components are an  important sign of phrase \textbf {conventionalization}. It seems that such phrases are stored as single units in the human memory.  In our case such conventionalized phrases included:  \begin{otherlanguage*}{russian}программное обеспечение\end{otherlanguage*} (\textit {software program}), \begin{otherlanguage*}{russian}земельный участок\end{otherlanguage*} (\textit {landing lot}), \begin{otherlanguage*}{russian}квадратный метр\end{otherlanguage*} (\textit {square meter}),   \begin{otherlanguage*}{russian}электронная почта\end{otherlanguage*} (\textit {electronic mail}), \begin{otherlanguage*}{russian}заработная плата\end{otherlanguage*} (\textit {employee wages}),   \begin{otherlanguage*}{russian}мобильный телефон\end{otherlanguage*} (\textit {mobile phone}),  \begin{otherlanguage*}{russian}лента новостей\end{otherlanguage*} (\textit {news feed}), and  \begin{otherlanguage*}{russian}торговый центр\end{otherlanguage*} (\textit {shopping center}).

Besides, we found  that the  average level of entropy (4.07) of phrase associations is much higher for non-thesaurus phrases than for thesaurus phrases (3.24). This means that associations of thesaurus phrases are more concentrated, more motivated by the phrase.  But at the same time some clearly compositional non-thesaurus phrases also have fairly low entropy of associations, for example, \begin{otherlanguage*}{russian}пресс-служба администрации \end{otherlanguage*} (\textit {press-service of the administration}).

\begin{table}[h]
\begin {scriptsize}
\begin{center}
\begin{tabular}{|l|c|c|c|c|c|}
\hline \bf Phrase &  \bf A12&\bf A21&\bf Ph1&\bf Ph2&\bf  Entr\\ \hline
\begin{otherlanguage*}{russian}финал лиги\end{otherlanguage*}& 0  &0&\bf 2&\bf 18&2.16\\ 
(league final) &   &&&&\\\hline
\begin{otherlanguage*}{russian}начало года\end{otherlanguage*}&   0&0&1&0&2.66\\ 
 (beginning of the year) &   &&&&\\\hline
\begin{otherlanguage*}{russian}ежедневный обзор\end{otherlanguage*}&  1 &0&\bf 3&\bf 14&2.90\\ 
(daily review) &   &&&&\\\hline
\begin{otherlanguage*}{russian}пресс-служба адми-\end{otherlanguage*}&   && & &\\ 
\begin{otherlanguage*}{russian}нистрации (press- \end{otherlanguage*}&  0 &0&\bf 14 &\bf 6 &2.97\\  
service of  administration)&   &&&&\\\hline
\begin{otherlanguage*}{russian}еженедельный обзор\end{otherlanguage*}&  0 &0&\bf 6&\bf 10&3.19\\
(weekly review) &   &&&&\\ \hline
\begin{otherlanguage*}{russian}необходимый доку-\end{otherlanguage*}&  1 &0&\bf 0&\bf 14&3.64\\
\begin{otherlanguage*}{russian}мент\end{otherlanguage*} (necessary document) &   &&&&\\ \hline
\begin{otherlanguage*}{russian}конец января\end{otherlanguage*}&  0 &0&\bf 2&\bf 7&3.81\\ 
(end of January) &   &&&&\\\hline
\begin{otherlanguage*}{russian}должность главы\end{otherlanguage*}&  0 &0&\bf 5&\bf 2&3.85\\ 
(post of the head) &   &&&&\\\hline\hline
\begin{otherlanguage*}{russian}новое поколение\end{otherlanguage*}&  0 &4&1&8&3.90\\
(new generation) &   &&&&\\ \hline
\begin{otherlanguage*}{russian}член совета\end{otherlanguage*}& 5  &0&2&5&3.96\\ 
(member of council) &   &&&&\\\hline
\begin{otherlanguage*}{russian}повышение \end{otherlanguage*}&  &&&&\\ 
\begin{otherlanguage*}{russian}эффективности\end{otherlanguage*}&0   &1&6&6&3.98\\ (increase in efficiency) &  &&&&\\\hline
\begin{otherlanguage*}{russian}увеличение объема\end{otherlanguage*}& 3  &0&7&4&4.02\\ 
(growth in volume) &   &&&&\\\hline
\begin{otherlanguage*}{russian}крупный размер\end{otherlanguage*}& 4  &0&6&4&4.07\\ 
(large size) &   &&&&\\\hline
\begin{otherlanguage*}{russian}миллион евро \end{otherlanguage*}&  0 &0&5&4&4.11\\ 
(million of euros) &   &&&&\\\hline
...&&&&&\\\hline
\begin{otherlanguage*}{russian}особое внимание\end{otherlanguage*}&  0 &2&0&5&4.63\\ 
(special attention)&  &&&&\\ \hline
\begin{otherlanguage*}{russian}председатель коми-\end{otherlanguage*}&  5 &0&5&3&4.63\\ 
\begin{otherlanguage*}{russian}тета\end{otherlanguage*}(chairman of committee) &  &&&&\\ \hline
\begin{otherlanguage*}{russian}экономический форум \end{otherlanguage*}& 0  &2&2&3&4.65\\ 
(economic forum)&  &&&&\\ \hline
\begin{otherlanguage*}{russian}интересный \end{otherlanguage*}&&&&&\\
\begin{otherlanguage*}{russian}комментарий\end{otherlanguage*}&0  &0&2&6&4.69\\ 
(interesting comment)&&&&&\\\hline
Average&\bf 1.22   &\bf 0.36&\bf 2.80&\bf 6.36&\bf 4.07\\ \hline

\end{tabular}
\end{center}
\end {scriptsize}
\caption{\label{font-table} Results of association experiments for non-thesaurus phrases }
\end{table}

We can also see that the phrases differ in the number of intersections between the associations obtained for a phrase and for its components. It seems natural that the already found conventionalized phrases have numerous  intersections of this kind (Table 2) because the phrase and its components are closely related to each other.

 On the contrary, other thesaurus phrases have a relatively small number of  such intersections. It means that the thesaurus phrases evoke their own associations more often. For example, the phrase \begin{otherlanguage*}{russian}повышение цен\end{otherlanguage*} (\textit {increase of prices}) has  frequent associations with the words \begin{otherlanguage*}{russian}инфляция\end{otherlanguage*} (\textit {inflation}) (16 of 25) and \begin{otherlanguage*}{russian}кризис\end{otherlanguage*} (\textit {crisis}), which were not mentioned as associations for its component words. On average, intersection between associations of the phrase and its component associations for non-thesaurus phrases is  four times less than for thesaurus phrases.

It   can also be seen that non-thesaurus phrases with low entropy of associations can have large numbers of intersections between the component associations and the phrase associations. In such cases, low entropy of the phrase associations is mainly detemined by its components, for example, their probable syntactic dependencies. Only one  of the non-thesaurus phrases has both low entropy of phrase associations and a few number of  intersections  of the phrase and component associations at the same time: \begin{otherlanguage*}{russian}начало года\end{otherlanguage*} (beginning of the year). It is highly associated with calendar months: \textit {January} and \textit {September}. For thesaurus phrases, a relatively high number of intersections between the phrase and component associations was revealed for most arguable thesaurus phrases: \begin{otherlanguage*}{russian}транспортное происшествие\end{otherlanguage*} (\textit {transport accident}) and \begin{otherlanguage*}{russian}температура воздуха\end{otherlanguage*} (\textit {air temperature}).

Thus, we can suppose that if a phrase has a low level of entropy of associations  together with a small number of the same associations for the phrase and its components then it is also conventionalized.  

We can introduce the threshold as 0.8*MaxEntropy of answers. MaxEntropy is the maximal entropy we can obtain if respondents give equiprobable  answers. In the current experiment, the threshold is equal to 3.76 for thesaurus phrases and 3.89 for non-thesaurus phrases. In our experiment, such conventionalized phrases include \begin{otherlanguage*}{russian}учебное заведение\end{otherlanguage*} (\textit {educational institute}), \begin{otherlanguage*}{russian}повышение цен\end{otherlanguage*} (\textit {increase in prices}), \begin{otherlanguage*}{russian}дорожное движение\end{otherlanguage*} (\textit {road traffic}), 
\begin{otherlanguage*}{russian}главный герой\end{otherlanguage*} (\textit {main hero}), \begin{otherlanguage*}{russian}медицинская помощь\end{otherlanguage*} (\textit {medical aid}).

A a result, we can say that we have found two signs of phrase conventionalization in the  association experiment described: 
\begin {itemize}
\item component words are frequently associated with each other, and
\item  associations of a phrase have both low entropy (less than 0.8*MaxEntropy) and a low level of intersection between component and phrase associations  (less than 20\%).
\end {itemize}

Using all three factors (association of component words to each other, entropy of phrase associations, and intersection of component word associations and phrase associations), it is possible to differentiate thesaurus phrases and non-thesaurus phrases with greater than 94\% accuracy. 

It is interesting to compare current results with the smaller amounts of associations. With this aim, we took the first 15 associations obtained for single words and phrases. The same above-mentioned thesaurus phrases have frequent mutual associations between components (that is, have association idiosyncrasy).

Phrases \begin{otherlanguage*}{russian}медицинская помощь\end{otherlanguage*} (\textit {medical aid}) and \begin{otherlanguage*}{russian}температура воздуха\end{otherlanguage*} (\textit {air temperature}) had entropy of associations more than 0.8*MaxEntropy. Only two non-thesaurus phrases had both low entropy (less than 0.8*MaxEntropy) and the low level of  intersection between assotiations of the phrase and its components: \begin{otherlanguage*}{russian}финал лиги\end{otherlanguage*} 
(\textit {league final}) and
\begin{otherlanguage*}{russian}начало года\end{otherlanguage*} (\textit {beginning of the year}). As a result, in this smaller experiment, the obtained associations can distinguish thesaurus phrases with accuracy more than 92\%.

\section{Detecting the Conventionalized Expressions with Distributional Models}

We  compared the results of the association experiment with the results of distributional models. In previous works, it was  supposed  that non-compositional phrases can be distinguished with comparison of the phrase distributional vector and distributional vectors of their components: it was supposed that the similarity is less for non-compositional phrases \cite {Cordeiro2016,Gharbieh}.

We used a Russian news collection (0.45 B tokens) and generated phrase and word embeddings with word2vec tool.  For the phrases under consideration, we calculated cosine similarity between the phrase vector $v(w_1w_2)$ and the sum of normalized vectors of phrase components $v(w_1+w_2)$ according to formula from \cite {Cordeiro2016}.

\[v(w_1+w_2)=(\frac{v(w_1)}{|v(w_1)|}+\frac{v(w_2)}{|v(w_2)|})\]

To evaluate different parameter sets,  we located all phrases in the ascending order of similarity scores. We wanted to check if
the  thesaurus phrases with idiosynrasy obtain lesser values of word2vec similarity than non-thesaurus phrases without any specificity. We utilized MAP (mean average precision measure) to evaluate the quality of ordering. 

We experimented with different parameters of word2vec and evaluated them with MAP on our data. We found that the best word2vec model (200 dimensions, 3 word window size) achieved quite low value of MAP ( \textbf{0.391}), which means that it is very difficult for current embedding models to differentiate  thesaurus and non-thesaurus phrases in our experiment. 

We can also calculate MAP for the same phrase list ordered accoring to the increased entropy of phrase associations. And here we obtain MAP equal to \textbf{0.642}. Thus, entropy of human associations without accounting additional factors predicts thesaurus phrases significantly better than the embedding models.

\section{Related Work}
The annotation of multiword expressions on compositionality/non-compositionality of noun compounds has been studied in several works \cite {Cordeiro,Reddy2011,Ramisch}.

Reddy et al. \shortcite {Reddy2011} created the set of 90 noun compounds. The phrases were taken  from WordNet.  For each compound, the following types of tasks 
have been given: a judgement on how literal the phrase is and a judgement on how literal each noun is within the compound. 
They used 30 turkers to obtain judgements on the compound compositionality in each task.

 Ramisch et al. \shortcite {Ramisch}  asked respondents about the degree to which the meaning of an expression follows from its components:    separately from each component and from both  components in total. The authors of the paper stress that such indirect annotation provides reliable and stable data. However, this approach was confronted with difficulties concerning the inconsistency of the answers in some cases. 
For example, English  speakers  agreed  on  the  level of head and head + modifier compositionality for  phrase \textit {dirty word},  but  disagreed  when  judging  the  modifier:  it  was  fully idiomatic for some, but others thought that the phrase just contained an uncommon sense of \textit {dirty}.

Maziarz et al. \shortcite {Maziarz} try to formulate the procedural definition of   multiword lexical units that should be included in the Polish wordnet so that
   lexicographers could apply these principles consistently.  Then they asked  linguists to classify phrases using this definition into three categories: \textit {multiword lexical unit}, \textit {not multiword lexical unit}, and \textit{don't know}. They concluded that  a group of 5-7 linguists is able to decide whether multiword lexical units should be introduced in a wordnet with the appropriate agreement. However,  this approach was considered  too expensive.

In another experiment, Maziarz et al. \shortcite {Maziarz} directed linguists to answer questions based on non-compositionality criteria of phrases including metaphoric character, hyponymy toward the syntactic head, ability to be paraphrased, non-separability, fixed word order, terminological register, etc. Then the answers were used to train the decision tree algorithm to predict inclusion or non-inclusion of an expression into the Polish wordnet. However, the obtained decision trees were different for the various phrase sets under analysis.

Farahmand  et al. \shortcite {Farahmand} describe the annotation of non-compositionality  and conventionalization  of noun compounds. They asked the annotators to
make  binary  decisions  about  compositionality of phrases. Compositional compounds were further annotated as conventionalized
 or non-conventionalized. A compound was considered as conventionalized in neither of its constituents can be substituted
for their near-synonyms. Sometimes the decision was diffucult because such phrases could really exist (\textit {floor space} vs. \textit {floor area}).
                  
To annotate the compounds, five experts were hired. In such a way, the authors \cite {Farahmand}
tried to avoid problems with crowdsourcing, which can lead to flaws in the results \cite {Reddy2011}.
The authors stress that  identifying conventionalization is not a trivial task and that human agreement on this property can be quite low. 
The examples of found compositional, but conventionalized phrases included: \textit {cable car, food court, speed limit}, etc.
The task of this study to distinguish conventionalized or non-conventionalized phrases among compositional compounds is the closest to our work.

For Russian there are two large resources of human associations. The well-known Russian Association dictionary \cite{Karaulov} is currently obsolete. Another assoicaion-oriented project Sociation.org\footnote{http://sociation.org/} collected a lot of current Russian associations but it does not have associations for the phrases under analysis.

Practical conclusions from the above-described experiments and related work are as follows:
\begin {itemize}
\item In annotating compositionality/non-compositionality of multiword expressions  by crowdsourcing  as in  \cite{Cordeiro,Reddy2011,Ramisch}, it is also  useful to ask respondents  about their associations for the phrase and its components to detect relational idiosyncrasy,
\item In expert analysis of multiword expressions for inclusion into computational resources as in \cite {Maziarz,Farahmand}, it is useful to ask experts about additional lexical or conceptual  relations that the phrase have and that do not follow from the phrase components,
\item In computational approaches of extracting non-compositional multiword expressions, it is useful to compare contexts of phrase occurrence and contexts of its component word occurrences trying to detect \textit {weirdness} in the  phrase context. 

\end {itemize}

\section{Conclusion}

In this paper, we have shown that if we want to obtain human evidence about  conventionalization of some phrases, we can ask native speakers about associations they have for a phrase and its component words. 

We have found that there are two forms of manifesting conventionalized phrases.  First, we can consider that a phrase is conventionalized if its component words have frequent associations to each other. The second type of conventionalized phrases can be revealed on the basis of two factors: low entropy of phrase associations and  a low number of intersections between component word and phrase associations. These three factors allows predicting conventionalized phrases with high accuracy. We have also shown that the existing embedding models  distinguish conventionalized phrases from non-conventionalized significantly worse.

In our opinion, developers of thesauri should consider the relational specificity (idiosynrasy) of multiword expressions, which can help them to decide on inclusion of specific phrases into their resources. Weird word co-occurrences with the phrase in comparison with its component contexts can be considered as an additional factor to detect conventionalized expressions in computational approaches.

\subsubsection*{Acknowledgments.} This study is supported by Russian Scientific Foundation (project N16-18-02074).  

\bibliography{mwe2017}

\begin{thebibliography}{}
\expandafter\ifx\csname natexlab\endcsname\relax\def\natexlab#1{#1}\fi

\bibitem[{Agirre et~al.(2006)Agirre, Aldezabal, and Pociello}]{Agirre}
Eneko Agirre, Izaskun Aldezabal, and Eli Pociello. 2006.
\newblock Lexicalization and multiword expressions in the basque wordnet.
\newblock In {\em Proceedings of Third International WordNet Conference\/}.
  pages 131--138.

\bibitem[{{ANSI/NISO}(2005)}]{Z39}
{ANSI/NISO}. 2005.
\newblock {\em Z39.19. Guidelines for the Construction, Format and Management
  of Monolingual Thesauri\/}.
\newblock ANSI/NISO.

\bibitem[{Baldwin and Kim(2010)}]{Baldwin}
Timothy Baldwin and Su~Nam Kim. 2010.
\newblock Multiword expressions.
\newblock In {\em Handbook of Natural Language Processing, Second Edition\/},
  Chapman and Hall/CRC, pages 267--292.

\bibitem[{Bentivogli and Pianta(2004)}]{Bentivogli}
Luisa Bentivogli and Emanuele Pianta. 2004.
\newblock Extending wordnet with syntagmatic information.
\newblock In {\em Proceedings of second global WordNet conference\/}. pages
  47--53.

\bibitem[{Bonial et~al.(2014)Bonial, Green, Preciado, and Palmer}]{Bonial}
Claire Bonial, Meredith Green, Jenette Preciado, and Martha Palmer. 2014.
\newblock An approach to take multi-word expressions.
\newblock In {\em Proc. of the 10th Workshop on Multiword Expressions\/}. pages
  94--98.

\bibitem[{Calzolari et~al.(2002)Calzolari, Fillmore, Grishman, Ide, Lenci,
  MacLeod, and Zampolli}]{Calzolari}
Nicoletta Calzolari, Charles~J Fillmore, Ralph Grishman, Nancy Ide, Alessandro
  Lenci, Catherine MacLeod, and Antonio Zampolli. 2002.
\newblock Towards best practice for multiword expressions in computational
  lexicons.
\newblock In {\em Proceedings of LREC-2002\/}.

\bibitem[{Cordeiro et~al.(2016{\natexlab{a}})Cordeiro, Ramisch, Idiart, and
  Villavicencio}]{Cordeiro2016}
Silvio Cordeiro, Carlos Ramisch, Marco Idiart, and Aline Villavicencio.
  2016{\natexlab{a}}.
\newblock Predicting the compositionality of nominal compounds: Giving word
  embeddings a hard time.
\newblock In {\em Proceedings of the 54th Annual Meeting of the Association for
  Computational Linguistics (Volume 1: Long Papers)\/}. Association for
  Computational Linguistics, pages 1986--1997.

\bibitem[{Cordeiro et~al.(2016{\natexlab{b}})Cordeiro, Ramisch, and
  Villavicencio}]{Cordeiro}
Silvio Cordeiro, Carlos Ramisch, and Aline Villavicencio. 2016{\natexlab{b}}.
\newblock Filtering and measuring the intrinsic quality of human
  compositionality judgments.
\newblock In {\em ACL 2016\/}. pages 32--37.

\bibitem[{Farahmand and Henderson(2016)}]{Farahmand1}
Meghdad Farahmand and James Henderson. 2016.
\newblock Modeling the non-substitutability of multiword expressions with
  distributional semantics and a log-linear model.
\newblock In {\em Proceedings of the 12th Workshop on Multiword Expressions,
  ACL 2016\/}. pages 61--66.

\bibitem[{Farahmand et~al.(2015)Farahmand, Smith, and Nivre}]{Farahmand}
Meghdad Farahmand, Aaron Smith, and Joakim Nivre. 2015.
\newblock A multiword expression data set: Annotating non-compositionality and
  conventionalization for english noun compounds.
\newblock In {\em Proceedings of NAACL-HLT\/}. Association for Computational
  Linguistics, pages 29--33.

\bibitem[{Gelbukh and Kolesnikova(2014)}]{Gelbukh}
Alexander Gelbukh and Olga Kolesnikova. 2014.
\newblock Multiword expressions in nlp: General survey and a special case of
  verb-noun constructions.
\newblock {\em Computational Linguistics: Concepts, Methodologies, Tools, and
  Applications,\/} pages 178--197.

\bibitem[{Gharbieh et~al.(2016)Gharbieh, Bhavsar, and Cook}]{Gharbieh}
Waseem Gharbieh, Virendra~C Bhavsar, and Paul Cook. 2016.
\newblock A word embedding approach to identifying verb--noun idiomatic
  combinations pages 112--118.

\bibitem[{Karaulov et~al.(1994)Karaulov, Sorokin, Tarasov, Ufimtseva, and
  Cherkasova}]{Karaulov}
Yuri Karaulov, Yu. Sorokin, E.~Tarasov, N.~Ufimtseva, and G.~Cherkasova. 1994.
\newblock {\em Russian Association Dictionary\/}.

\bibitem[{Loukachevitch and Dobrov(2014)}]{Loukachevitch2014}
Natalia Loukachevitch and Boris Dobrov. 2014.
\newblock Ruthes linguistic ontology vs. russian wordnets.
\newblock In {\em Proceedings of Global WordNet Conference GWC-2014\/}. pages
  154--162.

\bibitem[{Loukachevitch and Lashevich(2016)}]{Loukachevitch2016}
Natalia Loukachevitch and German Lashevich. 2016.
\newblock Multiword expressions in russian thesauri ruthes and ruwordnet.
\newblock In {\em Proceedings of the AINL FRUCT 2016\/}. FRUCT, pages 66--71.

\bibitem[{Maziarz et~al.(2015)Maziarz, Szpakowicz, and Piasecki}]{Maziarz}
Marek Maziarz, Stan Szpakowicz, and Maciej Piasecki. 2015.
\newblock A procedural definition of multi-word lexical units.
\newblock In {\em Proceedings of Recent Advances in NLP Conference
  RANLP-2015\/}. pages 427--435.

\bibitem[{Mel'{\v{c}}uk(2012)}]{Melchuk}
Igor Mel'{\v{c}}uk. 2012.
\newblock Phraseology in the language, in the dictionary, and in the computer.
\newblock {\em Yearbook of Phraseology\/} 3(1):31--56.

\bibitem[{Miller(1998)}]{Miller}
George~A Miller. 1998.
\newblock Nouns in wordnet.
\newblock {\em WordNet: An electronic lexical database\/} pages 24--45.

\bibitem[{Pearce(2001)}]{Pearce}
Darren Pearce. 2001.
\newblock Synonymy in collocation extraction.
\newblock In {\em Proceedings of the workshop on WordNet and other lexical
  resources, second meeting of the north american chapter of the association
  for computational linguistics\/}. pages 41--46.

\bibitem[{Pecina(2010)}]{Pecina}
Pavel Pecina. 2010.
\newblock Lexical association measures and collocation extraction.
\newblock {\em Language resources and evaluation\/} 44(1-2):137--158.

\bibitem[{Piasecki et~al.(2015)Piasecki, Wendelberger, and Maziarz}]{Piasecki}
Maciej Piasecki, Michal Wendelberger, and Marek Maziarz. 2015.
\newblock Extraction of the multi-word lexical units in the perspective of the
  wordnet expansion.
\newblock In {\em RANLP-2015\/}. pages 512--520.

\bibitem[{Ramisch et~al.(2016)Ramisch, Cordeiro, Zilio, Idiart, Villavicencio,
  and Wilkens}]{Ramisch}
Carlos Ramisch, Silvio Cordeiro, Leonardo Zilio, Marco Idiart, Aline
  Villavicencio, and Rodrigo Wilkens. 2016.
\newblock How naked is the naked truth? a multilingual lexicon of nominal
  compound compositionality.
\newblock In {\em Proceedings of the 54th Annual Meeting of the Association for
  Computational Linguistics (Volume 2: Short Papers)\/}. Association for
  Computational Linguistics, pages 114--133.

\bibitem[{Reddy et~al.(2011)Reddy, McCarthy, and Manandhar}]{Reddy2011}
Siva Reddy, Diana McCarthy, and Suresh Manandhar. 2011.
\newblock An empirical study on compositionality in compound nouns.
\newblock In {\em IJCNLP\/}. pages 210--218.

\bibitem[{Sag et~al.(2002)Sag, Baldwin, Bond, Copestake, and Flickinger}]{Sag}
Ivan~A Sag, Timothy Baldwin, Francis Bond, Ann Copestake, and Dan Flickinger.
  2002.
\newblock Multiword expressions: A pain in the neck for nlp.
\newblock In {\em Proceedings of International Conference on Intelligent Text
  Processing and Computational Linguistics, CICLING-2002\/}. Springer Berlin
  Heidelberg, pages 1--15.

\bibitem[{Senaldi et~al.(2016)Senaldi, Lebani, and Lenci}]{Senaldi}
Marco~SG Senaldi, Gianluca~E Lebani, and Alessandro Lenci. 2016.
\newblock Lexical variability and compositionality: Investigating idiomaticity
  with distributional semantic models pages 21--31.

\bibitem[{Vincze and Almasi(2014)}]{Vincze}
Veronika Vincze and Attila Almasi. 2014.
\newblock Non-lexicalized concepts in wordnets: A case study of english and
  hungarian.
\newblock In {\em Proceedings of Global WordNet, Conference GWC-2014\/}. Global
  WordNet Association.

\end{thebibliography}
\bibliographystyle{acl_natbib}

\end{document}